\pdfoutput=1
\documentclass[11pt]{article}

\usepackage[final]{acl}
\usepackage{xcolor}
\usepackage{todonotes}
\usepackage{amsmath}
\usepackage{xurl}
\usepackage{amssymb}
\usepackage{times}
\usepackage{latexsym}
\usepackage{float}
\usepackage{pifont}
\usepackage{enumitem}
\usepackage{booktabs}
\newcommand{\cmark}{\ding{51}}%
\usepackage[T1]{fontenc}
\usepackage{caption}

\usepackage[utf8]{inputenc}

\usepackage{microtype}

\usepackage{inconsolata}

\title{On the Rigour of Scientific Writing: Criteria, Analysis, and Insights}
\author{\textbf{Joseph James$^1$\footnotemark[1], Chenghao Xiao$^2$\footnotemark[1], Yucheng Li$^3$, Chenghua Lin$^{1,4}$\footnotemark[2]}\\
    $^1$Department of Computer Science, The University of Sheffield, UK\\
    $^2$Department of Computer Science, Durham University, UK\\
    $^3$Department of Computer Science, University of Surrey, UK\\
    $^4$Department of Computer Science, The University of Manchester, UK\\
    \texttt{jhfjames1@sheffield.ac.uk} \hfill \texttt{chenghao.xiao@durham.ac.uk} \\
    \texttt{yucheng.li@surrey.ac.uk} \hfill \texttt{chenghua.lin@manchester.ac.uk}}

\begin{document}
\interfootnotelinepenalty=10000 
\maketitle

\renewcommand{\thefootnote}{\fnsymbol{footnote}}
\footnotetext[1]{Equal contribution.}
\footnotetext[2]{Corresponding author.}
\renewcommand*{\thefootnote}{\arabic{footnote}}

\begin{abstract}
Rigour is crucial for scientific research as it ensures the reproducibility and validity of results and findings. 
Despite its importance, little work exists on modelling rigour computationally, and there is a lack of analysis on whether these criteria can effectively signal or measure the rigour of scientific papers in practice. 
In this paper, we introduce a bottom-up, data-driven framework to automatically identify and define rigour criteria and assess their relevance in scientific writing. Our framework includes rigour keyword extraction, detailed rigour definition generation, and salient criteria identification. Furthermore, our framework is domain-agnostic and can be tailored to the evaluation of scientific rigour for different areas, accommodating the distinct salient criteria across fields. 
We conducted comprehensive experiments based on datasets collected from two high impact venues for Machine Learning and NLP (i.e., ICLR and ACL) to demonstrate the effectiveness of our framework in modelling rigour. 
In addition, we analyse linguistic patterns of rigour, revealing that framing certainty is crucial for enhancing the perception of scientific rigour, while suggestion certainty and probability uncertainty diminish it. 
%Experimental results on ICLR, ACL, and REF datasets, along with human evaluations, demonstrate the effectiveness of our framework.

\end{abstract}

\section{Introduction}

%\textcolor{red}{[CL: (1) something on reproducibility -- ER proposal and my proposal; (2) lack of condenses on the definition of rigour; (3) need to provide some discussion on how existing work analyse rigour; (4) compared to novelty, rigour is poses more challenge for analysis]}

Rigour is one of the cornerstones of scientific research. Despite its profound importance and the widespread use of the term in both scientific and lay parlance, the scientific literature adds surprisingly little to our understanding of rigour, with the term almost always used without a definition, as if its meaning is self-evident. There are few definitions for scientific rigour available. For instance, the National Institutes of Health (NIH) has defined scientific rigour as ``\textit{the strict application of the scientific method to ensure robust and unbiased experimental design, methodology, analysis, interpretation and reporting of results. This includes full transparency in reporting experimental details so that others may reproduce and extend the findings.}'' \cite{nih}. 
%With an extended checklist for animal related research\footnote{\url{https://arriveguidelines.org/arrive-guidelines}}. 
%The level of granularity achieved in the definition and checklist, first published in 2010 and fully refined in 2023, represents a significant advancement in rigour in the field. For a well-established field, this time frame for such refinement would be considered relatively brief. However for a relatively new and ever changing field such as Computer Science, this rate of development may not be fast enough to keep pace with the constant changes and devopment in the field.
Whilst this definition may seem useful, it has been criticised for being both overly verbose and disconcertingly vague \cite{casadevall2016rigorous}.
%\footnote{\url{https://www.aclweb.org/portal/content/efficient-nlp-policy-document}}.

Inherently, scientific rigour is multi-faceted---no single criterion can define it fully. Indeed, there are some obvious dimensions of rigour, such as reproducibility, as rigorous scientific practice can enhance the likelihood that the results generated will be reproducible. 
%Unfortunately, there exists a phenomenon known as the `reproducibility crisis' in science \cite{baker20161} – with 70\% of scientists failing to reproduce someone else’s results at least once, and more than half failing to reproduce their own results. 
In Computer Science (CS), recent years have witnessed an exponential increase in the number of publications (e.g., for AI, Machine Learning, and NLP), culminating in nearly half a million publications worldwide in 2021 alone \cite{maslej2023artificial}. 
This surge has resulted in what is termed \textit{scientific debt}, where researchers prioritise `novel' methods without sufficiently grounding their work in theory, conducting extensive ablation studies, or performing comprehensive evaluations \cite{nityasya2023scientific}. Furthermore, this trend has exacerbated the \textit{reproducibility crisis} in science \cite{baker20161},  a widespread problem where many scientific studies are difficult or impossible to reproduce by other researchers. 
%This surge has led to what is termed 'scientific debt,' where researchers prioritize 'novel' methods without sufficiently grounding their work in theory, conducting extensive ablation studies, or performing comprehensive evaluations  \cite{nityasya2023scientific}, as well as a phenomenon known as the `reproducibility crisis' in science \cite{baker20161} – with 70\% of researchers failing to reproduce someone else’s results at least once, and more than half failing to reproduce their own results. 

The community has begun to address some of the issues surrounding scientific rigour. For instance, in response to calls for more transparent and robust research, ACL introduced separate scores for soundness and excitement in 2023\footnote{\url{https://2023.aclweb.org/blog/overall-recommendation/}}, and ICLR now includes a breakdown for correctness, technical and empirical novelty, and significance in its review process\footnote{\url{https://iclr.cc/Conferences/2022/ReviewerGuide}}.
While these initiatives are important, there is a significant gap: 
%a lack of consensus on the criteria for rigour, 
\textit{different research domains may have varying preferences and traditions for defining rigour criteria, and there is a lack of analysis on whether those criteria can effectively signal or measure the rigour of scientific papers in practice}.   
Efforts have been made to model different aspects of scientific writing. Notably, there is a rich literature on modelling scientific discourse, which aims not merely to present information and ideas but also to ensure their effective communication, allowing readers to accurately perceive what authors intend \cite{goldsack2023domain}. 
Additionally, there are studies focusing on detecting scientific novelty from text \cite{savov2021measuring,luo2022combination}, as novelty is regarded as an important aspect of judging scientific merit. However, compared to novelty, rigour is more challenging to define and analyse, and little work exists on modelling and analysing rigour computationally, particularly in the domain of CS. The limited existing work often focuses on non-CS domains (e.g., biomedical research), employs a top-down approach  to defining criteria for scientific rigour (e.g., based on researchers’ own experiences), or lacks empirical analysis on the effectiveness of the defined rigour dimensions \cite{prager2019improving}.

To address the aforementioned gap in modelling scientific rigour, we have developed a \textit{bottom-up}, \textit{data-driven} framework that can automatically elicit candidate rigour criteria and construct their detailed definitions. 
Furthermore, our framework is domain-agnostic and can support the analysis of the salience of criteria for signalling the rigour of scientific papers--an important analysis which, to the best of our knowledge, has not been attempted previously. 
Specifically, we 
(i) construct a high-quality corpus of publications that exemplify high rigour, where the corpus is used to train a binary rigour classifier by fine-tuning SciBERT \cite{beltagy-etal-2019-scibert}. We then extract candidate rigour keywords from two much larger datasets, namely ICLR\footnote{\url{https://github.com/berenslab/iclr-dataset}} and ACL Anthology\footnote{\url{https://github.com/shauryr/ACL-anthology-corpus}}, based on the predictions of the rigour classifier using feature selection. Next, (ii) we generate detailed definitions for the candidate rigour keywords by prompting GPT-4, and (iii) analyse the salience of criteria for signalling the rigour of scientific papers by proposing an LLM-based embedding approach.

%\textcolor{orange}{Specifically, our framework consists of three main components: (i) extraction of rigour keywords (or aspects?) using feature selection from datasets collected from ACL and ICLR (see §\ref{sec:dataset} for details). The initial set of rigour keywords also underwent manual inspection for quality control; (ii) generation of detailed definitions for the extracted rigour aspects by prompting GPT-4; and (iii) identification of salient criteria for signalling or measuring the rigour of scientific writing. This is achieved by leveraging LLM embeddings to analyse the distributional differences in categorical variables between rigorous and less rigorous papers.}

% 
% From the paper we classify the abstract and introduction using Longformer SciBERT \cite{beltagy-etal-2019-scibert}. We focus on the abstract and introduction due to the nature of these sections being consistent over multiple different conferences and domains allowing us to compare papers from multiple different topics in Computer Science. 

%We conducted experiments on three datasets (i.e., ICLR, ACL, and REF). To measure the salience of rigour criteria, we enumerated all possible combinations of extracted rigour criteria, and computed the similarity between the respective rigor criteria using GritLM \cite{muennighoff2024generative} to quantify the difference in similarity distribution between the predicted rigour level and similarity score to identify the salient criteria.
We conducted comprehensive experiments based on datasets collected from two high impact conferences for ML and NLP (i.e. ICLR and ACL) to demonstrate the effectiveness of our framework in modelling rigour, involving analysing the performance of rigour classifier, and assessing the most salient rigour criteria for each experimented dataset quantitatively. 
We hypothesise that linguistic pattern differences exist between high rigour and less rigour papers, impacting readers' perception of scientific rigour. Therefore, we further conducted a sentence-level analysis based on the aspect-level uncertainty theory of \cite{pei-jurgens-2021-measuring}. Experimental results reveal that framing certainty is crucial for enhancing the perception of scientific rigour, while suggestion certainty and probability uncertainty diminish it. 
%We further analysed each sentence in both classes with each rigour criterion and calculated the aspect-level uncertainty \cite{pei-jurgens-2021-measuring}. We found that sentences characterised by high rigour exhibited different levels of certainty compared to those with low rigour. 
%Finally, we conducted human evaluations and observed higher agreement for sentences containing higher rigour papers.
%4* sentences in comparison to non-4* sentences.
To summarise, our contributions are three-fold:
\begin{itemize}[noitemsep,topsep=1pt,parsep=1pt,partopsep=1pt]
    \item We propose a bottom-up, domain-agnostic computational framework that automatically identifies candidate rigour criteria and generates their corresponding definitions.
    \item We create a set of Rigour Criteria and propose an LLM-embedding based method that can effectively measure the salience of specific rigour criteria for a given research domain.
    \item Our comprehensive analysis provides valuable insights into the linguistic features that signify perceived rigour in scientific writing, promoting transparency and robustness in scientific research.
\end{itemize}

\section{Related Work} \label{relatedwork}

%\subsection{Criterion of Scientific Rigour}\label{rigourwork}
%\noindent\textbf{Criterion of Scientific Rigour.} 
\subsection{Criterion of Scientific Rigour}
%Science advances by sharing results with others; therefore, it is crucial to ensure rigour in scientific research to make sure that studies lead to reliable, unbiased knowledge and reduce uncertainty in reproduction \cite{hofseth2018getting}. Many funding agencies, such as the Research Excellence Framework (REF) in the UK and the National Institutes of Health (NIH) in the US, consider rigour as an essential factor in their assessment and allocation of public research funding \cite{REF_2021_Guidance,nih}.
Despite its fundamental importance, 
%there doesn't seem to be a universal criterion for scientific rigour. 
existing guidelines or definitions for rigour are often vague and general, such as the NIH's suggestion to justify the methodology, identify potential weaknesses, and address limitations \cite{johnson2020review,wilson2021three}. \citet{sansbury2022rigor} highlight the importance of rigour in study design and conduct, statistical procedures, data preparation, and availability. In addition, there exist many domain-specific requirements for rigour proposed by researchers. For example, \citet{lithgow2017long} believe stricter variability control is necessary for animal research, 
%such as controlling lifespan differences in worms, using shared batches of reagents,
following strict handling protocols, and adhering to precise methodological guidelines. \citet{hamberg1994scientific} discuss the multifaceted nature of rigour and suggest that different criteria should be used to assess truthful findings under different circumstances. In family medicine research, \citet{hamberg1994scientific} argue that researchers should focus on credibility, dependability, confirmability, and transferability instead of traditional rigour criteria.
%such as validity, reliability, and objectivity. 
In the computer science community, leading conferences like NeurIPS and ACL have employed a checklist approach for authors to self-review their submissions and address issues of research ethics and reproducibility \cite{neurips,arr}. This encourages authors to clearly describe their research questions, explicitly explain the limitations of their work, and report experiments in as much detail as possible.

These aforementioned criteria are predominantly developed in a top-down fashion, relying heavily on domain experts' experience. Such an approach has several limitations in real practice. Firstly, due to the multi-faceted nature of rigour, it may be challenging to directly apply existing rigour criteria or checklists from other domains. This makes it difficult to scale and adapt rigour assessment practices across various disciplines. Secondly, the criterion/checklist approach has a limited impact on the actual reviewing process. 
For instance, authors are only required to complete a checklist which can challenge reviewers trying to assess the level of rigour as additional details are often not required to support the claims.

%For instance, authors may provide insufficient level of detail when filling out the checklist, and reviewers may find it challenging to judge the extent of rigour for submissions as normally all checklist items were ticked by authors. 
Moreover, \citet{Randall_2023} suggests that even professional reviewers might excessively focus on novelty while neglecting the importance of rigour. These highlight the potential and need for bottom-up computational modelling of rigour, where such models can assist authors in improving their narrative and writing, and help editors assess the rigour and truthfulness of paper submissions.

\subsection{Computational Modelling of Scientific Rigour}

Several attempts have been made to computationally analyse the rigour of scientific papers. For example, \citet{Wael} investigated how researchers use the word "rigour" in information system literature but discovered that the exact meaning was ambiguous in current research. Additionally, various automated tools have been proposed to assess the rigour of academic papers. \citet{phillips2017online} developed an online software that spots genetic errors in cancer papers, while \citet{sun2022assessing} used knowledge graphs to assess the credibility of papers based on meta-data such as publication venue, affiliation, and citations.  %However, these methods are neither domain specific nor provide substantial guidance for the authors to improve their narrative and writing. 
However, these methods are neither domain-specific nor do they provide sufficient guidance for authors to improve their narrative and writing.
In contrast, SciScore \cite{SciScore_2024} is an online system that uses language models to produce rigour reports for paper drafts, helping authors identify weaknesses in their presentation. However, they rely on existing rigour checklists suggested by NIH and MDAR \cite{chambers2019towards}, which are not easily scalable or transferable to other domains.

%\noindent\textbf{Research Excellence Framework (REF).}
\subsection{Research Excellence Framework (REF)}
%Many funding agencies, such as the Research Excellence Framework (REF) in the UK and the National Institutes of Health (NIH) in the US, consider rigour as an essential factor in their assessment and allocation of public research funding \cite{REF_2021_Guidance,nih}.
REF is the UK's national system for assessing the quality of research across UK universities. The assessment is carried out once every 7 years, and the assessment outcome informs the distribution of research funding nationwide, which constitutes a significant portion of each university’s research income. 
Due to its importance, all universities identify the strongest outputs for REF submission (e.g., if a researcher has multiple NeurIPS/ACL publications, only the strongest among them will be submitted, as there is a stringent and limited cap on the number of submissions allowed per individual). The submitted output will then undergo reviews by panel members consisting of senior academics, who will rate the paper using a five-point scale from 4*-\textit{Quality that is world-leading in terms of originality, significance and rigour}, to 0*-\textit{quality that falls below the standard of nationally recognised work}. 
In addition, 4* papers are given much heavier weights than papers from other categories, e.g., one 4* paper will be allocated \textit{four times} the funding of one 3* paper. 
See Appendix \ref{tab:REFstars} for a full description of the categories.

\section{Methodology}

%\textcolor{red}{[Some overarching text about the framework and the justification]} 
%\textcolor{green}{[Alot of the technicality in the experimental setup]} 

%There is a lack of consensus on the criterion for rigour and a lack of analysis of the real perception of rigour by researchers. 

\begin{figure*}[t]
\centering
\includegraphics[width=0.9\linewidth]{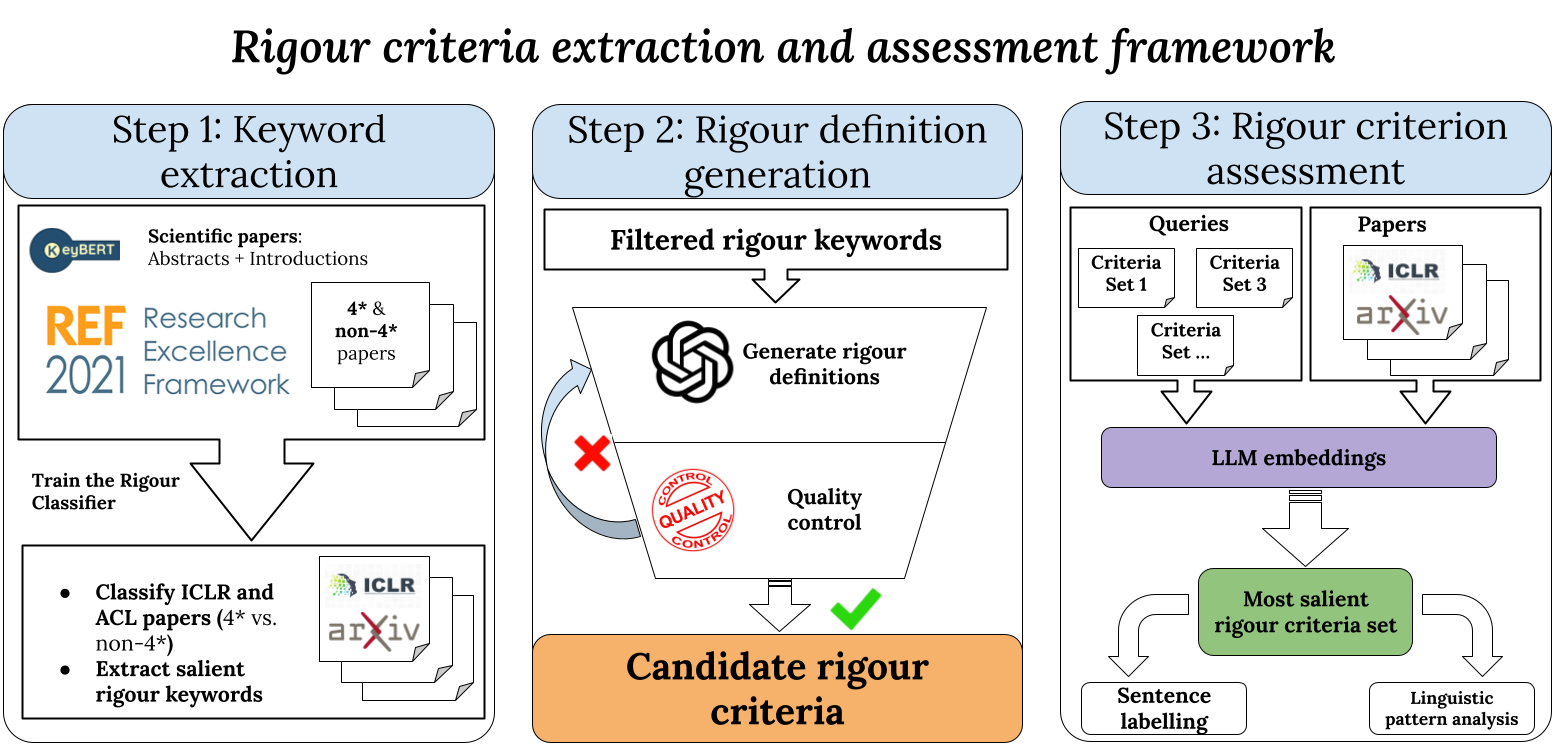}
\caption{Illustration of the rigour criteria extraction and assessment framework.}
\label{fig:model}
\end{figure*}

In this section, we describe our data-driven, bottom-up framework for eliciting the criterion for defining the rigour of scientific writing, which consists of three main components. An illustration of the framework is shown in Fig \ref{fig:model}.

% ----------- fig:framework  -----------

% ----------- end of fig -----------

%In this work, we have explored and analysed different ways to extract features from text without having finegrained labels. The goal of this approach is to see how we can extract features in a systematic way such that it can be used to understand scientific writing. We develop a step by step process that can distinguish rigour and present the results.

\subsection{Rigour Keyword Extraction}\label{sec:rigour_keyword}

Extracting candidate keywords that are highly relevant to the rigour of scientific papers requires a corpus of papers that exemplify high rigour. While it may seem reasonable to use the review scores of papers as an indicator of rigour (e.g., papers from ICLR with higher correctness scores are regarded as more rigorous compared to those with lower scores), we argue that this might not be a robust proxy.
%Prior research has highlighted inconsistencies in the reviewing process, noting that approximately half of the accepted papers would change if the review process was randomly rerun 
Prior research has highlighted inconsistencies in the review process, noting that the acceptance decisions for approximately half of the papers would change if the process were repeated
\cite{cortes2021inconsistency,beygelzimer2023has}. %\textcolor{orange}{Additionally, over a third of papers rejected from NeurIPS were eventually published in other high-quality venues \cite{beygelzimer2023has}.}

\noindent\textbf{High rigour paper source.}~~To address this issue, we opted for a more reliable source—papers classified under the Research Excellence Framework (REF). The rationale, as discussed in  §\ref{relatedwork}, is that the stringent nomination and review processes (i.e., by expert panel members) provide very high confidence that the 4* publications classified in the REF will exhibit high rigour, which is one of the main assessment criteria.
Since REF only publishes scores of publication outputs at the institutional level rather than for each individual publication, we collected publication outputs from institutions predominantly rated as 4* (e.g., Imperial College London and Oxford), as well as from institutions whose outputs are predominantly rated as non-4*.
With this, we constructed a REF dataset consisting of papers categorised into two groups: 4* and non-4* (please refer to Table \ref{tab:dataset} for statistics).

\noindent\textbf{Bias Mitigation.}~~ We performed preprocessing to minimise various types of biases that could affect the classification of rigour. 
%\textcolor{red}{[CL: provide some explanation here why removing bias is important for the task]}. 
First, we manually removed mandated keywords, footers, headers, and section titles from the papers and retained only the abstracts and introduction sections, instead of the entire articles. 
%There is an intuitive sense that rigour is more prevalent in other sections of a paper, such as the methods section, however,While other sections are also likely to reflect the quality or rigour of scientific writing, we focus on abstract and introduction due to several factors: (i) a prior study by \citet{afzal2020deep} performed rigour classification based on the title and abstract in the Biomedical domain, achieving satisfactory performance; (ii) the REF dataset contains papers from various disciplines within Computer Science, not just NLP and machine learning. As a result, we observe a lack of standardisation across journals and conferences in other sections, e.g., large variations in paper length and the number of sections, as well as the limitations section not always being present. Therefore, we decided to focus on the abstract and introduction, which provide a comprehensive overview of the paper (e.g., contextual foundation, motivation, research problem/objectives, methods and results/findings) and could mitigate potential biases (e.g., paper length and number of sections) in training due to the relatively small dataset size. 
%
due to the lack of standardised structure across various areas of computer science publications and journals \cite{posteguillo1999schematic}, we decided against the use of other sections to avoid adding bias (i.e. paper length and mandatory sections such as limitations in ACL papers that may be optional in other publications). \citet{afzal2020deep} showed that the title and abstract are enough to classify rigour in the Biomedical domain. Based on their findings, we believe the inclusion of the introduction section would further strengthen the classifier's performance.

Additionally, we removed information related to the authors, publication venues, and affiliated institutions to mitigate author and institution biases \cite{thelwall2022can2}. To mitigate topical biases (e.g., theoretical machine learning, HCI, medical, etc.), we utilise the method proposed by \citet{golchin2023not} and extracted domain-specific keywords using KeyBERT \cite{grootendorst2020keybert} then replaced the keywords with [MASK].
%Topic words in the papers were masked using KeyBERT \cite{grootendorst2020keybert} to mitigate topical biases (e.g. theoretical, medical, etc.), by replacing the keywords with the [MASK] token \cite{golchin2023not}. 
%Through this meticulous process, we curated a dataset that contains \textcolor{red}{988 abstracts and introductions}  that are of high in quality for classifying the rigour scientific papers.
Through this meticulous process, we curated a high-quality dataset containing the abstracts and introductions of 988 papers, which were used to train and test our rigour classifier. 
\newline\noindent\textbf{Keyword Extraction.}~~
We train a SciBERT \cite{beltagy-etal-2019-scibert} binary classifier for rigour on the constructed REF dataset. As our framework is designed to be domain-agnostic, we extract rigour keywords from two much larger datasets, namely ICLR and ACL (see Table \ref{tab:dataset} for statistics), which represent two subfields of computer science; machine learning and NLP. This is achieved by first predicting the labels (4* or non-4*) of the papers from the two datasets, followed by feature extraction using mutual information \cite{kraskov2004estimating} to separate out the important features for 4* and non-4*. Positive coefficients indicate a higher association with 4* papers, while negative coefficients indicate a higher association with non-4* papers. The keywords prominent in 4* papers are shown in Fig \ref{fig:reg_coef} and for non-4* papers in Appendix \ref{fig:neg_reg_coef}. The extracted candidate rigour keywords then underwent manual filtering for quality control. 
%keywords such as \textit{setting}, \textit{biases}, \textit{contribution} were selected based on prior work (as discussed in §\ref{relatedwork}), while keywords such as \textit{benchmarks} and \textit{generalise} were selected based on our domain knowledge in Computer Science.
%(Selecting keywords such as \textit{setting}, \textit{generalise}, \textit{contribution}, that can be associated with rigour),
We provide the full list of the top 100 keywords  in Appendix \ref{tab:features}.

\begin{table}[t]
\centering \small
\begin{tabular}{lcccccc}
\toprule
 & \textbf{4*} & \textbf{non-4*} & \textbf{Train} & \textbf{Val} & \textbf{Test} & \textbf{Total} \\
 \midrule

REF & 292 & 696& 790 &  99 & 99 & 988 \\
ICLR  & - & - & - & - & - & 5,493\\
ACL   & - & - & - & - & - & 32,651\\
\bottomrule
% Total  & 292 & 696 & 790 & 99 & 38243 & 38243\\
% \hline
\end{tabular}
\caption{\centering Dataset statistics for all three datasets}\label{tab:dataset}
\end{table}

%We utilise the contextual representations of SciBERT \cite{beltagy-etal-2019-scibert} to fine-tune our classifier. 
%Salient features are then extracted from the predicted ICLR and ACL abstracts and introductions. This approach enables the consideration of a broader and more generalised labelled dataset. We lemmatise and remove stopwords from the text and vectorise the words using TFIDF. The most informative textual features were extracted using either Chi-squared or mutual information feature selection techniques. To evaluate the predictive performance of these features, three different supervised learning algorithms were employed; Logistic Regression, Random Forest Classifier, and Support Vector Classification. The goal was to determine which classifier achieved the highest classification accuracy on the task. This allows only the most salient features to be extracted. Combined with previous work in rigour and the salient features, we select keywords manually to be used as potential rigour criteria,for example examples "setting","generalise" and "contribution". While other features are present, we aim to extract one dimension (rigour) to see whether it is enough to distinguish the two REF classes.

\subsection{Rigour Definition Generation}
Following the extraction of rigour keywords in  \S\ref{sec:rigour_keyword}, we generated definitions using GPT-4\footnote{Mixtral-8x7B, Gemini-1.5, and Claude-3 Opus were also tested and resulted in similar outputs.}
\cite{achiam2023gpt}, using the following prompt: 
``\textit{Give the definition of "[keyword]" in the context of Computer science and Machine learning. In the format: [keyword]: Refers to [definition]}''. 

% \noindent An example for the definition of \textbf{"baselines"} generated by GPT-4\footnote{gpt-4-turbo-2024-04-09} is as followed: 
% ``\textit{Baselines: refers to reference points or initial measurements that serve as a starting point for comparison or evaluation. Baselines provide a foundation for assessing the performance, effectiveness, or efficiency of systems, algorithms, models, or solutions.}'', demonstrating the reliability of LLM generated definitions. 

We validate our generated criteria manually to determine whether our outputs are adequate at explaining each rigour keyword (more details given in \S 5.2). For example, the definition of \textbf{reproducibility} by \citet{raghupathi2022reproducibility} is:  
``\textit{the ability of an independent research team to produce the same results using the same research method based on the documentation made by the original research team}'',
\noindent in comparison to our GPT-4\footnote{gpt-4-turbo-2024-04-09} generated definition: 
``\textit{Refers to the ability to reliably recreate the same results or outputs from a given model or experiment, given the same input data and configuration settings, by providing the complete source code and using openly available tools and datasets}''.
\noindent Overall, our manual examinations verify that the GPT-4 generated definitions are of good quality. A full list of generated definitions are shown in Appendix \ref{tab:definitions}.
%\textcolor{orange}{We then combined unique sets of rigour criteria (e.g., \textbf{\{Settings, Baselines\}}, \textbf{\{Examples, Benchmarks, Justifications\}}, etc), which will be used in \S\ref{sec:rigour_assessment} for determining which rigour criteria are more prevalent in 4* papers than in non-4* papers.} 
% By using the equation
% \begin{equation}
%    N  = \sum_{r \subseteq C} \frac{C!}{r!(C-r)!}
% \end{equation}
% where N is the total number of combinations, C is the total criteria set and r is the subset of elements in the criteria set.
\subsection{Salient Rigour Criterion Assessment} \label{sec:rigour_assessment}

To the best of our knowledge, no prior work has attempted to analyse the salience of criteria for signalling rigour. 
%Having identified rigour keywords and generating their definitions, 
%We leverage LLM-based embedding models to derive insights \textit{grounded in semantics}, regarding criteria that significantly distinguish 4* (rigorous) and non-4* (less rigorous) papers. 
We approach the problem by assessing the prominence of a rigour criterion or a set of rigour criteria (e.g., \textbf{\{Settings, Baselines\}}, \textbf{\{Examples, Benchmarks, Justifications\}}) based on their semantic similarity to 4* and non-4* papers. By doing so, we measure whether papers containing a specific set of rigour criteria associate more with 4* papers than non-4* papers. % to distinguish what features separate the two.
%\textcolor{orange}{We then combined unique sets of rigour criteria (e.g., \textbf{\{Settings, Baselines\}}, \textbf{\{Examples, Benchmarks, Justifications\}}, etc), which will be used in \S\ref{sec:rigour_assessment} for determining which rigour criteria are more prevalent in 4* papers than in non-4* papers.} 

\noindent\textbf{LLM Embeddings.} 
% Recent advancements in decoder-based embedding models \cite{wang2023improving,muennighoff2024generative} present great promise in achieving reasoning-level language understanding with embeddings \cite{xiao2024rar}. In essence, training embedding models with LLM backbones is aligning their representational abilities with their generative abilities \cite{muennighoff2024generative}. Concretely, these models are trained with a diverse array of class, InfoNCE loss \cite{oord2018representation}
We leverage LLM-based embedding models to calculate the semantic similarity between each rigour criterion definition and papers from the REF, ICLR and ACL datasets. 
The rationale behind opting for off-the-shelf LLM-based embedding models are three-fold: (i) the lack of fine-grained ground-truth labels for each criterion, thus unable to train a criterion-specific classifiers; (ii) given the limited size of the REF dataset, training a domain-specific embedding model with contrastive learning essentially enforces a uniform embedding space for this specific task, resulting in subpar performance; and (iii) recent advancements in LLM-based embedding models show exceptional generalisation, alignment on reasoning-level language, and instruction-following abilities \cite{wang2023improving,muennighoff2024generative,xiao2024rar}. Therefore, we argue that it is reasonable to judge rigour by encoding rigour criteria and scientific papers into the same semantic space, and conducting similarity matching to identify the prevalence of the rigour criteria for each paper. 
In our experiment, we specifically use the representations of GritLM \cite{muennighoff2024generative}, a 7B model that unifies generative and representational abilities in one model and achieves state-of-the-art results on MTEB (Massive Text Embedding Benchmark \cite{muennighoff2022mteb}) and RAR-b (Reasoning as Retrieval Benchmark \cite{xiao2024rar}). 

\noindent\textbf{Computing semantic similarity.} 
%With the following instruction: ``\textit{Given the following definitions, retrieve the appropriate document that contains the following criteria:}'', we appended the query (combination of the criteria with its associated definition) to the instruction.
We prepend the query (combinations of the criteria with its associated definition) with the following instruction: ``\textit{Given the following definitions, retrieve the appropriate document that contains the following criteria:}''.
The concatenated instruction and query are passed to the model, and we apply mean pooling to the query tokens, giving us $E(q|i)$ (i.e. query embeddings conditioned on the instruction). Aligning with common practices in instruction-aware embedding systems \cite{asai2022task,su2022one}, we encode the documents without instructions, giving the document embedding $E(d)$.
The cosine similarity between $E(q|i)$ and $E(d)$ is taken as the indicator of a document reaching the corresponding criterion $q$, allowing our analysis on the corpus-level distributional difference. Formally, we have 
\begin{equation}
\mathbf{E}(q|i) = \frac{1}{|q|} \sum_{t \in q} \mathbf{E}([i; q])_t; \mathbf{E}(d) = \frac{1}{|d|} \sum_{t \in d} \mathbf{E}(d)_t
\end{equation}
where $i$ denotes the instruction built upon the rigour criteria, \( \mathbf{E}([i; q])_t \) denotes the embedding of the \( t \)-th token in the concatenated sequence \( [i; q] \), and \( |q| \) is the number of tokens in the query \( q \); \( \mathbf{E}(d)_t \) denotes the embedding of the \( t \)-th token in the document \( d \), and \( |d| \) is the number of tokens in document \( d \) without prepending instructions. And 
$\text{cos\_sim}(\mathbf{E}(q|i), \mathbf{E}(d)) = \frac{\mathbf{E}(q|i) \cdot \mathbf{E}(d)}{\|\mathbf{E}(q|i)\| \|\mathbf{E}(d)\|}$ is taken as the indicator of document $d$ meeting rigour criteria $i$. 
%\textcolor{blue}{Finally, we can further identify the most salient set of rigour criteria by aggregating the semantic similarities between the criteria set and the 4* and non-4* papers.} 

Finally, we identify the most salient set of rigour criteria by analysing whether a significant distributional difference exists by comparing the semantic similarities between the queried set of criteria and the 4* and non-4* papers. Notably, we discovered that appending sets of criteria to a single query is more effective at separating the semantic similarities in comparison to individual criterion queried then summed, as shown in Appendix \ref{fig:query}. 

\section{Experimental Setup}

%\paragraph{Datasets} \label{sec:dataset}
\noindent\textbf{Datasets.}
The REF dataset was constructed based on REF 2021\footnote{\url{https://2021.ref.ac.uk/}}, which covers papers published between 2014 and 2021. The submissions from UOA (Unit of Assessment) 11 \footnote{\url{https://results2021.ref.ac.uk/profiles/units-of-assessment/11}}, which covers all areas of computer science and information, was used to create the dataset. As described in \S\ref{sec:rigour_keyword}, we collected publications from institutions whose outputs are predominantly rated as 4* (e.g., Imperial College London and Oxford), as well as from institutions whose outputs are predominantly rated as non-4*, to form a binary labelled dataset for rigour. 

For both the ICLR and ACL papers, we extracted abstracts and introductions from existing datasets or from arXiv\footnote{\url{https://www.kaggle.com/datasets/Cornell-University/arxiv/data}}. For ICLR, we considered all papers from 2022 and 2023 \cite{gonzalez2024learning}, to show the effectiveness of our rigour classifier on papers submitted after 2021. The ACL dataset was developed using the ACL Anthology Corpus \cite{acl_anthology_corpus}. Given the high formatting consistency within their respective domains, we automatically extracted abstracts and introductions from both datasets using pattern matching with section titles. The statistics of the three datasets are presented in Table~\ref{tab:dataset}.

\noindent\textbf{Rigour classifier and feature extraction.}~~We trained the rigour classifier by fine-tuning a Longformer version of SciBERT\footnote{\url{https://huggingface.co/yorko/scibert_scivocab_uncased_long_4096}}, a pre-trained language model trained on scientific publications that has proven effective at capturing language patterns and domain-specific knowledge within scientific texts \cite{beltagy-etal-2019-scibert}.  
%Longformer \cite{Beltagy2020Longformer} is a Transfomer-based models which are able to process long sequences of text. By conducting experiments on SciBERT Longformer we are able to capture domain-specific features for the abstract and introduction. 
The full settings, including the model hyperparameters, are given in Appendix~\ref{tab:hyperparameters}.
For feature extraction, we implemented mutual information with logistic regression. The default parameters provided by the scikit-learn library\footnote{\url{https://scikit-learn.org/stable/}} were used.

\noindent\textbf{GritLM-7B.}~~We used the embedding-only variant of the GritLM-7B model \cite{muennighoff2024generative}, and the representations without the LM head. Mean pooling is taken over token-level embeddings to attain sentence-level embeddings. 
%We concatenated the instruction to just the queries and not the documents \cite{asai2022task,su2022one}.
For query embeddings, the final attention is only given to actual query tokens conditioned by the prepended instructions \cite{muennighoff2024generative}, and for document embeddings, we simply take the mean pooling of all tokens.
% cosine similarity is calculated from the query and instruction against the documents (abstracts and introductions). 

% We label sentences by calculating the cosine similarity score for each rigour criteria against all sentences and assigning the criteria with the highest similarity score to each individual sentence. The instruction used with the query is as followed ``\texttt{Given the following definition, retrieve the sentence that aligns with the definition:}''.

%\noindent\textbf{Evaluation metrics}~~Due to the robustness to outliers and non-parametric nature of the metric, we use Kendall correlation \cite{kendall1938new} in order to demonstrate the correlation of our rigour criteria similarity score with REF labels. The values range from [-1,1] where the value of -1 shows strong negative correlation and 1 shows a strong positive correlation. 

\section{Experimental Results}

\noindent\textbf{5.1~~Rigour classification and feature extraction} \\
Experimental results of rigour classification based on the REF dataset demonstrated that our classifier gives strong performance, signalling that the abstract and introduction of a paper can provide rich information for predicting the rigour of papers.  
%\textcolor{orange}{We note that the introduction provides good indication signals in terms of what criteria are present throughout the paper, even if limited in detail, and this is reflected in our classifiers' accuracy in predicting rigour. [CL TODO: move this to the result section]}
More specifically, our classifier trained on our processed data achieved an accuracy of 0.94 and F1 score of 0.90, while the unmasked data (topic words included) resulted in an accuracy of 0.93 and F1 score of 0.88. The results highlight the robustness of the classifier in distinguishing 4* and non-4* papers. 

\begin{figure*}[t]
\centering
\includegraphics[width=1\textwidth]{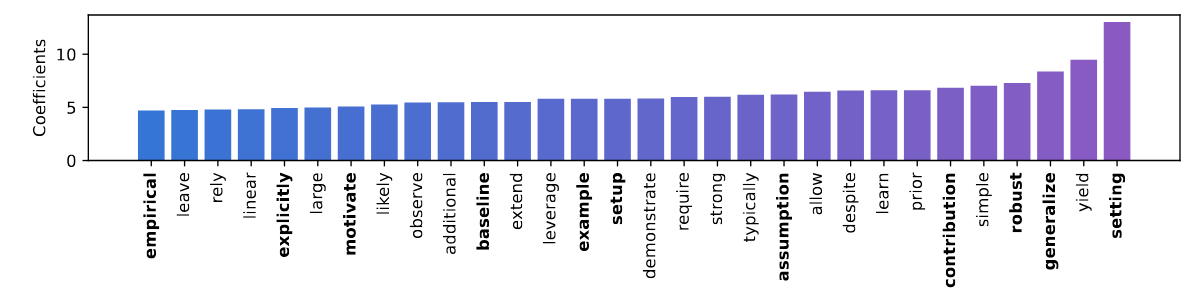}
\caption{Top 30 salient keywords for 4* predicted papers from ICLR and ACL using Mutual Information with candidate rigour keywords highlighted in \textbf{bold}. List of keywords in Table~\ref{tab:features}.}
\label{fig:reg_coef}
\end{figure*}

We then predicted the rigour labels for the ICLR and ACL dataset with our classifier, and performed feature selection based on mutual information to identify the most salient keywords. The top 30 salient keywords for 4* papers are shown in Fig~\ref{fig:reg_coef}, with keywords related to rigour highlighted. For examples \textit{"Setting"}, \textit{"Generalise"} and \textit{"Robust"}.
%The results of our classifiers indicates that the abstract and introduction can accurately classify rigour labels, highlighting the potential to analyse the difference in rigour between 4* and non-4* papers. 
 %This shows that the model learned on more general features prominent across all papers.

%We justify our classifier by showing a shift in percentage 4* predicted and ICLR paper correctness scores, shown in Table \ref{tab:correctness}. This result falls inline with evidence from \cite{thelwall2023can}, 
%stating agreement between reviewers for scientific quality is only moderate. 
%and discussed in \S\ref{sec:rigour_keyword}. This draws attention to the need for a unified set of rigour criteria across peer reviews to allow for a more consistent and standardised reviewing process.

% \begin{table}[t]
% \centering \small 
% \begin{tabular}{l|c}
% \hline
% Correctness & predicted rigour \\
% \hline
% 4 &  0.8181\\
% 3 &  0.7697\\
% 2 &  0.7288\\
% 1 & 0.6713\\
% \hline
% \end{tabular}
% \caption{ICLR correctness score vs percentage of 4* predicted abstracts and introductions}\label{tab:correctness}
% \end{table}

% \begin{table}[t]
% \centering \small 
% \begin{tabular}{l|c}
% \hline
% Acceptance & predicted rigour \\
% \hline
% Oral &  0.8408\\
% Spotlight &  0.8321\\
% Poster &  0.8004\\
% Reject & 0.7290\\
% \hline
% \end{tabular}
% \caption{ICLR acceptance score vs rigour percentage}\label{tab:acceptance}
% \end{table}

\noindent\textbf{5.2~~Assessing the Salience of Rigour criteria} \\  
From the salient keywords, we manually selected 13 potential rigour criteria. An additional three criteria were chosen based off of prior work, these include \textit{"Reproducibility"}, \textit{"limitations"} and \textit{"justifications"}, derived from literature discussed in §\ref{relatedwork}. Following this, we generated 65,535 unique combinations of the criteria to be used to query REF, ICLR and ACL papers, based on the unique permutations of criteria $C$ and subset of elements in the criteria set $r$, given by $N  = \sum_{r \subseteq C} \frac{C!}{r!(C-r)!}$.

\begin{figure*}[tp]
\centering
\includegraphics[width=1\textwidth]{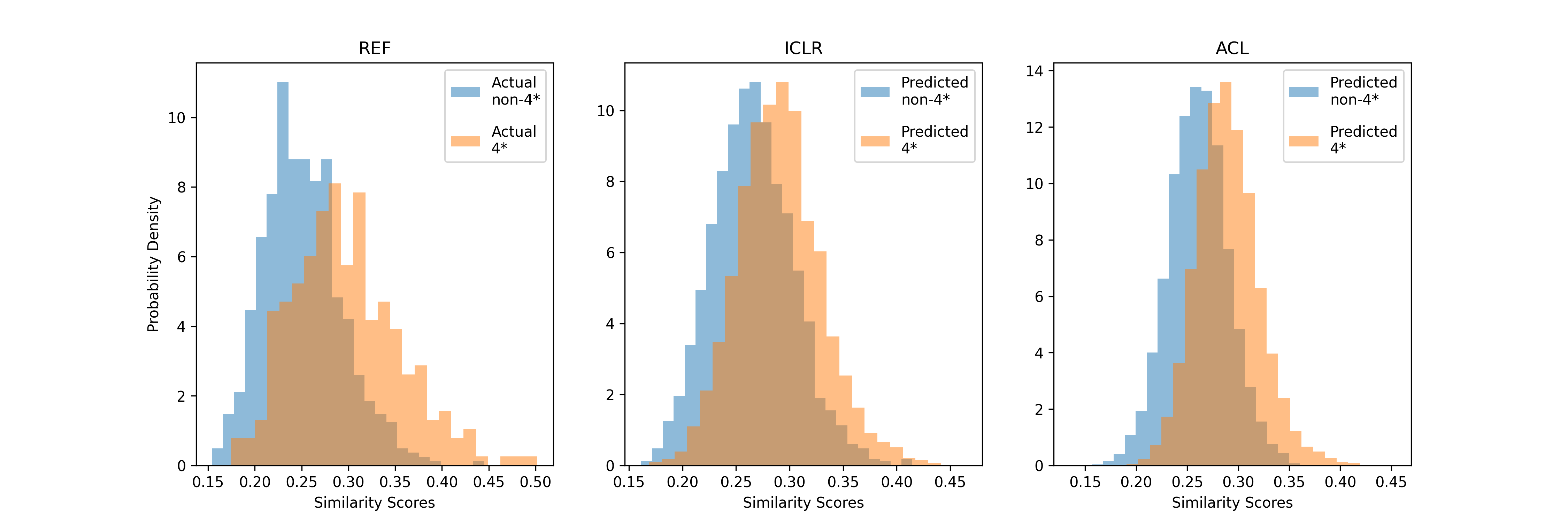}
\caption{Distribution of similarity score for best rigour criteria set for each of REF, ICLR and ACL datasets}
\label{fig:rig_plot}
\end{figure*}

The most salient rigour criteria set, i.e., criteria with the highest correlation with 4* papers, are presented in Table \ref{tab:rig_criteria}, with the corresponding distributions of the cosine similarity between the rigour criteria and each document split by 4* and non-4*, shown in Fig \ref{fig:rig_plot}. The cosine similarity score here is calculated via the documents and their corresponding salient rigour criteria set.
The cosine similarity values indicate the prominence of the rigour criteria set which we utilise to signify what makes 4* papers different to non-4* papers. The correlation between rigour labels and the cosine similarity score are 0.307 for REF, 0.227 for ICLR and 0.240 for ACL ($p< 0.0001$). 

%A moderate correlation was observed for REF, which is stronger than for ICLR and ACL. 

A moderate correlation was observed for REF, while a weaker correlation was observed for ICLR and ACL. Nonetheless, we can observe a clear difference in distribution from Fig \ref{fig:rig_plot}, indicating 4* papers exhibit higher similarity scores with our salient rigour criteria set. By using Kendall's correlations, we justify the statistical significance of our correlations \cite{gilpin1993table}.

%By using Kendall's correlations, a more robust correlation metric compared to Pearson's r, we justify the statistical significance of our correlation \cite{gilpin1993table}.

The results in Table \ref{tab:rig_criteria} suggest each dataset has a slightly different rigour criteria set preference, indicating inherent domain-specific characteristics.  
%even within the respective domain.
We observe that criteria \textit{"Baselines"}, \textit{"Benchmarks"}, \textit{"Assumptions"} and \textit{"Reproducibility"} are prominent across all three datasets. This falls inline with recent work emphasising the importance of reproducibility \cite{semmelrock2023reproducibility}. On the other hand, \textit{"Challenges"} and \textit{"Contributions"} show little difference between the 4* and non-4* papers. This suggests that all papers contain these criteria to a similar extent, indicating a consensus across all domains.
In summary, we reveal that different sub-fields exhibit some degree of rigour criteria set preference, which can be captured by our framework in a data-driven way.

%the evaluation of scientific rigour for different areas to accommodate the different prominence of criteria across datasets.

% , indicates the need for a systematic framework like ours to tailor the evaluation of scientific rigour of different areas.

% Our analysis highlights that, While textual information contained within the abstract and introduction correlates with rigours, it is insufficient for a comprehensive assessment. Full evaluation of rigour requires the integration of external metadata and domain expertise \cite{thelwall2022can}. However, our findings reveal specific criteria can be used to enhance the communication of rigour. For instance, emphasising reproducibility and the use of benchmarks can significantly contribute to the perceived rigour of research publications.

\begin{table}[t]
%\captionsetup{width=0.5\textwidth}
\small
\centering
\begin{tabular}{lccc} 
\toprule
\textbf{Criteria} & \textbf{REF} & \textbf{ICLR} & \textbf{ACL} \\
\midrule
Biases & \textbf{--} & \cmark &  \cmark \\
Settings & \textbf{--} & \cmark & \cmark \\
Constraints & \cmark & \cmark & \textbf{--}\\
Limitations & \textbf{--} & \textbf{--} & \cmark \\
Baselines & \cmark & \cmark &  \cmark\\
Benchmarks & \cmark & \cmark & \cmark \\
Empirical Findings & \cmark & \cmark & \textbf{--}\\
Examples & \cmark & \textbf{--} & \textbf{--} \\
Motivations & \textbf{--} & \textbf{--} & \cmark \\
Generalisation & \cmark & \textbf{--} & \cmark \\
Robustness & \textbf{--} & \cmark & \cmark \\
Assumptions & \cmark & \cmark & \cmark \\
Justifications & \cmark & \cmark & \textbf{--} \\
Challenges & \textbf{--} & \textbf{--} & \textbf{--} \\
Contributions & \textbf{--} & \textbf{--} & \textbf{--} \\
Reproducibility & \cmark & \cmark & \cmark \\
\bottomrule
\end{tabular}
\caption{The most salient rigour criteria sets for the REF, ICLR, and ACL datasets, where \cmark \ indicates that a particular criterion is included in the criteria set for a specific dataset, and \textbf{--} \ indicates its absence. Essentially, each column corresponds to a criteria set.}
%\caption{Rigour criteria for REF, ICLR and ACL. Where \cmark \ indicates if the criteria is prominent and \textbf{--} \ indicates no difference.}
\label{tab:rig_criteria}
\end{table}

\noindent\textbf{5.3~~Linguistic Patterns of Scientific Rigour} \\
We first separate the papers into sentences for both  4* and non-4* papers. Then each sentence is labelled via GritLM with a rigour criterion. We utilise a threshold of 0.5 for the cosine similarity between rigour criterion and sentences to remove sentences that are not similar enough to any of the criterion. This is to exclude sentences such as prior work, which are prevalent in the papers but irrelevant to our rigour criteria.
% mislabelling of sentences such as sentences related to prior work, which is not part of our rigour criteria but is prevalent in the papers.
% We labelled sentences using GritLM by converting the abstracts and introductions into sentences. 
% Then \textcolor{red}{utilising the rigour criterion prominent} in at least one of the datasets (excluding "Challenges" and "Contributions"), we matched sentences with the highest cosine similarity to a criterion with a threshold of 0.5. 
%\textcolor{red}{We use sentences from ACL and ICLR so that we can compare in-domain differences of sentences in Machine Learning.}
% Our analysis only relies on the ACL and ICLR dataset, which ensures the topic consistency of sentences.
In total, we obtain 400k sentence labels which associated with a rigour criteria, a full breakdown of labels are shown in the Appendix \ref{tab:breakdown_labels}. We utilise these sentences to extract linguistic patterns that highlight the differences between 4* and non-4* sentences.

\noindent\textbf{Certainty metric.} 
Certainty is a crucial concept in scientific writing when communicating knowledge to the reader, especially for conveying rigour \cite{national2017communicating}. For example \textit{framing certainty} indicates how \textit{certain} or \textit{confident} scientific findings are ``framed'' and ``interpreted''.
Here we evaluate certainty of different aspect for the rigour criteria we identified previously.
Specifically, we use the certainty classifier introduced in \cite{pei-jurgens-2021-measuring} to label sentences with a certainty aspect.
We do this for both 4* and non-4* papers, which allows us to find the most relevant aspect of certainty for each rigour criterion in 4* or non-4* sentences.

% To evaluate certainty, we calculated the distribution of aspect-level certainty for each rigour criteria for both classes (4* and non-4*). Then the difference in distribution is calculated for each certainty aspect. This allows us to identify which certainty aspect is more prominent in 4* or non-4* sentences for each rigour criteria.

\begin{figure}[t]
\centering
\captionsetup{width=0.44\textwidth}
\includegraphics[width=.48\textwidth,height=0.33\textheight]{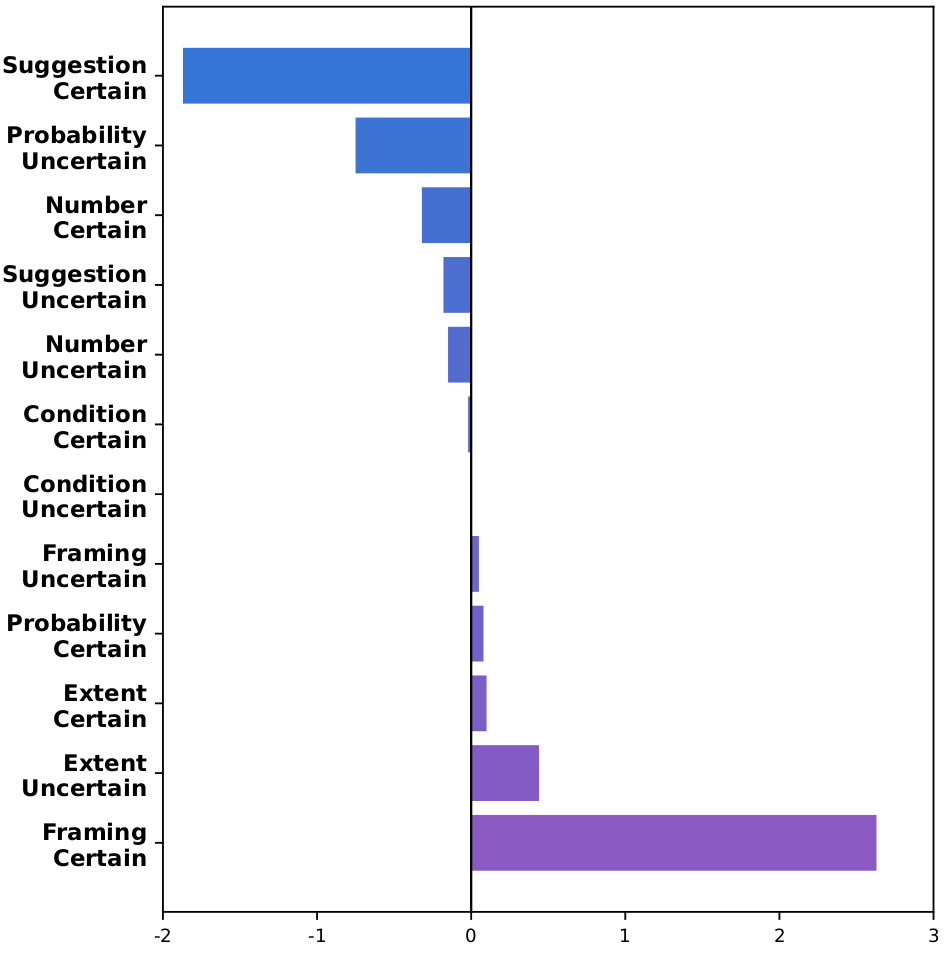}
\caption{Uncertainty scores for rigour criteria present in both ICLR and ACL. Positive values indicate 4* preference while negative values indicate non-4* preference.}
\label{fig:certainty}
\end{figure}

Sentences with rigour criteria contained in both ACL and ICLR (Biases, settings, Baselines, Benchmarks, Robustness, Assumptions and Reproducibility), 346k in total, were used in our evaluation and the most relevant aspects of certainty were identified by their differences in probabilities between 4* and non-4* sentences, as shown in 
Fig \ref{fig:certainty}. It can be observed that 4* sentences are more likely to contain \textit{framing certainty}.
% , i.e., how \textit{certain} or \textit{confident} scientific findings are ``framed'' and ``interpreted''.
% . Framing certainty refers to how scientific research is interpreted and perceived by readers and aims to make the findings more objective. 
% This is a commonly seen more often in scientific abstracts than in news articles \cite{pei-jurgens-2021-measuring}. 
But for non-4* sentences, \textit{suggestion certainty} and \textit{probability uncertainty} are the more common phenomenons, where \textit{suggestion certainty} refers to how certain the findings propose suggestions or future actions, and \textit{probability uncertainty} indicates the usage of uncertain wording such as ``possibly''.
These findings on linguistic patterns around certainty highlight the importance of scientific writing, as it has direct impact on readers' perception of the rigour of a paper.
% This implies that the perceived impact of the research may not be as crucial for indicating scientific rigour. 
We provide examples in Appendix \ref{rigour_examples} and  further analysis on all sentences in Appendix \ref{app:aspect_breakdown}.

\begin{table}[t] 
\centering \small
\begin{tabular}{p{0.10\textwidth}p{0.30\textwidth}}
\toprule
\textbf{Certainty} & \textbf{Example}  \\
\midrule
Framing  \quad certainty  & The most widespread family of techniques \textbf{are} diagnostic models, which use the internal activations of neural networks trained on a particular task as input to another predictive model. \\
\\

Suggestion certainty  & Such systems \textbf{need} to be assessed by system developers for any possible technological improvements and novel research ideas and by potential users for quality comparison purposes. \\
\\

Probability uncertainty & Various features can \textit{potentially} be used, based on the source and target context as well as syntactic and semantic analysis. \\
\\
\bottomrule

\end{tabular}
\caption{A sample of sentences from the three prevelant certainty classes. Words highlight in \textbf{bold} indiciate the certainty and words in \textit{italics} indicates the uncertainty.}\label{tab:example_certain}
\end{table}

%For the "Benchmark" criterion, 4-star sentences demonstrate a 2.09\% higher Framing certainty, while non 4-star sentences demonstrate a 3.71\% higher Suggestion certainty. This indicates that sentences with non-4-star ratings are more likely to discuss benchmarks in the context of implications or future actions for the scientific community. On the other hand, 4-star sentences tend to focus on how benchmarks are interpreted and framed. Full breakdown of criterion against aspect level certainity is shown in the Appendix \ref{tab:certainty}.

\begin{table}[t]
\centering 
\begin{tabular}{lcc}
\toprule
\textbf{Label} & \textbf{Rigour rating} & \textbf{Cohen's kappa} \\
\midrule
4*  & 3.4143 & \textbf{0.349}\\
non-4* & 3.3714 & 0.030\\
\bottomrule
\end{tabular}
\caption{Human evaluation on 35 pairs of sentences from rigour criteria contained in both ICLR and ACL.}\label{tab:human_eval}
\end{table}

\noindent\textbf{Human evaluation.}  We further conducted human evaluation to determine if the perceived level of rigour was different in 4* and non-4* sentences. We recruited two CS postgraduate students to evaluate their preference from pairs of sentences from rigour criterion contained in both ICLR and ACL. We selected 35 examples which was calculated using power analysis, this allowed the estimation of the smallest samples required for human evaluation to be statistically significant (for Cohen's d of 0.57, statistical power of 0.80 and significance level of 0.05 \cite{schuff2023human}). The evaluators were given guidelines to act as a reviewer assessing sentence pairs in isolation along with the associated rigour criterion. The guideline for rigour includes: (i) is the presented sentence written confidently about the criterion (i.e., confidence and certainty); (ii) does the sentence contain enough information to interpret the criterion (i.e., level of detail and relevance); and (iii) is the sentence straight to the point without excessive filler words (i.e., conciseness). 

To mitigate the potential confounding effect of sentence-level topic differences within each rigour criterion, we ensured that sentence pairs we annotated were semantically similar, verified with SciBERT using cosine similarity.
% we utilise SciBERT to retrieve the most semantically similar sentence pairs from both the REF classes. 
This allowed us to assess the nuanced differences in the sentences, eliminating the potential impact of different topics and findings. Sentences were assessed in pairs to give evaluators a frame of reference when scoring the level of perceived rigour and presented in a random anonymised order. Evaluators were given a 5-point Likert scale to rate both sentences separately. 

As shown in Table \ref{tab:human_eval}, we observe that the inter-annotator agreement for 4* sentences is fair while for non-4* sentences there was no agreement, demonstrating the consensus on sentences from 4* papers and signifying a higher level of perceived rigour. This is further backed up by our certainty results in Fig \ref{fig:certainty}, as specific certainty aspects are favoured in 4* papers such as \textit{framing certainty}. The relatively low agreement scores is not surprising due to rigour being a highly complex and abstractive concept. However, the key take-away of the results is the higher agreement of more rigorous sentences (from 4* papers) compared to less rigorous sentences (from non-4* papers). This shows more rigour in a sentence leads to less ambiguity and thus more likely to be agreed upon by the evaluators. We provide examples of 4* and non-4* sentences with respect to the rigour criteria in Appendix \ref{rigour_examples}.

% \noindent\textbf{Importance of communication.} From our findings, we observe common rigour criteria across different conferences within Computer Science. While universal criteria may be difficult with the current state of research publication, our results indicate a promising path towards a unified standard. The criteria may not completely distinguish the difference between the \textcolor{red}{REF classes}, but being able to demonstrate a noticeable difference from textual information indicates that a preference exists that lead to papers being more highly regarded \cite{davis2012scientific} within the scientific community.

\section{Conclusion}

In this paper, we introduce a bottom-up, data-driven framework to automatically identify and define rigour criteria and assess their relevance in scientific writing. Our framework is domain-agnostic and can be tailored to the evaluation of scientific rigour for different areas. 
Comprehensive experiments based on datasets demonstrate the effectiveness of our framework in modelling rigour. 
In addition, we analyse linguistic patterns of rigour, revealing that framing certainty is crucial for enhancing the perception of scientific rigour, while suggestion certainty and probability uncertainty diminish it.

\section*{Limitation}

\noindent \textbf{LLMs.}~~ The use of Large language models as a definition generator, semantic measure and sentence label annotator has its limitations, this is due to the stochastic nature of such models that may not capture all the nuances in the text compared to expert annotators. We utilised an embedding-based variant to partially address this limitation in the semantic measure, though it is not without its own drawbacks.

\noindent \textbf{Domain specific.}~~ Our approach looks at Computer Science papers and more specifically Machine Learning. Extending to other domains would lead to a more generalised rigour criteria, however this may not be desirable due to differences across domains which would reduce the descriptiveness of the criteria. 

\noindent \textbf{Full paper classification.}~~ Given the variability in formatting and structure across publications, we focused the analysis on the abstracts and introductions of the papers. This approach allowed more consistent evaluation across topics. However, to develop a comprehensive understanding of the textual factors that contribute to rigour, it is crucial that future investigation should analyse other sections.

% \section{Bib\TeX{} Files}
\label{sec:bibtex}

\section*{Acknowledgements}
Joseph James is supported by the Centre for Doctoral Training in Speech and Language Technologies (SLT) and their Applications funded by UK Research and Innovation [grant number EP/S023062/1].

\bibliography{custom}

\appendix
 \section{Appendix}
\label{sec:appendix}

\subsection{REF star rating description}\label{tab:REFstars}

\begin{table}[H]
\centering
\begin{tabular}{lp{0.3\textwidth}}
\toprule
\textbf{Quality level} & \textbf{Description} \\
\midrule
Four star & Quality that is world-leading in terms of originality, significance and rigour. \\
\\
Three star & Quality that is internationally excellent in terms of originality, significance and rigour but which falls short of the highest standards of excellence. \\
\\
Two star & Quality that is recognised internationally in terms of originality, significance and rigour \\
\\
One star & Quality that is recognised nationally in terms of originality, significance and rigour. \\
\\
Unclassified & Quality that falls below the standard of nationally recognised work. Or work which does not meet the published definition of research for the purposes of this assessment. \\

\bottomrule

\end{tabular}
\caption{REF star rating description \cite{REF_2021_Guidance}}
\end{table}

\subsection{Model training parameters}\label{tab:hyperparameters}

Model trained on two RTX4090's with hyperparameters shown in Table \ref{hyperparameters}.
\begin{table}[h]
\centering
\begin{tabular}{lc}
\toprule
\textbf{Setting} & \textbf{Value} \\
\midrule
num\_train\_epochs & 5 \\
per\_device\_train\_batch\_size & 1 \\
per\_device\_eval\_batch\_size & 1 \\
warmup\_steps & 100 \\
weight\_decay & 0.01 \\
learning\_rate & 5e-5 \\
logging\_steps & 50 \\
evaluation\_strategy & 'steps' \\
eval\_steps & 50 \\
load\_best\_model\_at\_end & True \\
metric\_for\_best\_model & 'f1' \\
gradient\_accumulation\_steps & 8 \\
seed & 41 \\
\bottomrule
\end{tabular}
\caption{\centering Model training hyperparameters}\label{hyperparameters}
\end{table}

\subsection{Key features for negative coefficients}\label{fig:neg_reg_coef}
\begin{figure*}[!hp]
    \centering
    \includegraphics[width=1\textwidth]{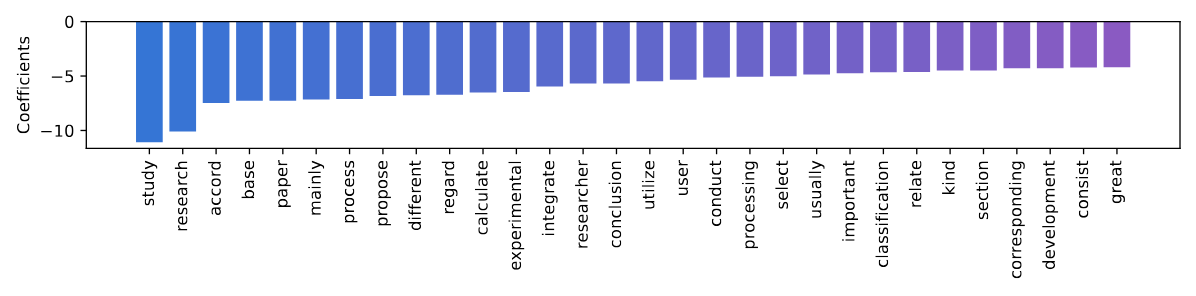}
    \caption{Top 30 salient keywords for non-4* predicted papers from ICLR and ACL using Mutual Information.}\label{top30_neg}
\end{figure*}

Top 30 Salient keywords for non-4* papers are shown in Fig \ref{top30_neg}.

\subsection{Top 100 Salient keywords}\label{tab:features}
The top 100 salient keywords for 4* papers are shown in Table \ref{100features}.
\begin{table*}[!hp]

\begin{tabular}{ccccc}
\toprule
\textbf{Chi-square features} \\
\midrule
\textbf{setting} & yield & \textbf{generalize} & \textbf{robust} & simple \\
\textbf{contribution} & prior & learn & despite & allow \\
\textbf{assumption} & typically & strong & require & demonstrate \\
\textbf{setup} & \textbf{example} & leverage & extend & \textbf{baseline} \\
additional & observe & likely & \textbf{motivate} & large \\
explicitly & linear & rely & leave & \textbf{empirical} \\
hold & estimate & \textbf{assume} & outperform & remove \\
instead & encode & replace & train & inference \\
tune & derive & produce & recent & gradient \\
subset & particular & contrast & remain & fail \\
directly & binary & unlike & property & consistent \\
natural & true & recently & distribution & fix \\
want & note & count & \textbf{constraint} & variable \\
denote & low & objective & let & log \\
align & embedding & downstream & state & right \\
parse & neural & optimization & non & \textbf{bias} \\
choose & simply & noise & small & average \\
differ & \textbf{benchmark} & deep & condition & unsupervised \\
draw & suggest & single & effect & like \\
fully & representation & perform & standard & high \\
\bottomrule
\end{tabular}
\caption{4* predicted rigour paper keywords extracted using Logistic regression and percentile Mutual information of 10\%, where bold text implies use as rigour criteria.}\label{100features}
\end{table*}

\subsection{Rigour criteria defintitions}\label{tab:definitions}
GPT-4 generated rigour criteria shown in Table \ref{tab:rigour_definitions}.

\begin{table*}[!hp]
\small\centering
\begin{tabular}{lp{0.8\textwidth}}
\toprule
\textbf{Criterion} & \textbf{Definition} \\
\bottomrule
Biases & Refers to systematic errors or distortions in the collection, analysis, or interpretation of data that can lead to skewed or inaccurate results. Uncertainties refer to the lack of precision or confidence in measurements, predictions, or conclusions due to limitations in data or knowledge. \\
\\
Settings & Refers to adjustable parameters or configurations that define the behavior or performance of a system or model. They are typically set before the learning or optimization process and impact the final outcome. \\
\\
Contributions & Refers to the original ideas, innovations, or advancements made by individuals or groups in the field. Contributions can take various forms and impact different aspects of computer science, including research, technology development, software engineering, algorithms, systems design, or theoretical advancements. \\
\\
Constraints & Refers to restrictions or bottlenecks imposed on a system, software application, algorithm, or problem-solving process. Constraints define the boundaries within which a solution or system must operate, and they help guide the design, implementation, and behaviour of computer systems. \\
\\
Limitations & Refers to the inherent shortcomings that exist within a system, technology, algorithm, or problem-solving approach. Limitations define the boundaries of what a system or solution can achieve or the constraints that restrict its performance, functionality, or applicability. \\
\\
Generalisation & Refers to the process of extracting common patterns, concepts, or properties from specific instances or examples and formulating more abstract or generalised representations or models. Generalisations help capture the essential characteristics or behaviours shared by a set of objects, data, or systems, enabling more efficient and flexible problem-solving, analysis, or design. \\
\\
Robustness & Refers to the ability of a system, software application, algorithm, or network to effectively handle and recover from abnormal or unexpected conditions, inputs, or events. A robust system is designed to withstand errors, exceptions, invalid inputs, or challenging operating conditions and continue functioning correctly or gracefully degrade without catastrophic failures. \\
\\
Benchmarks & Refers to standardised tests, metrics, or reference points used to measure and evaluate the performance, efficiency, or capability of computer systems, software applications, algorithms, or hardware components. Benchmarks provide a basis for comparing different systems or solutions and assessing their relative strengths and weaknesses. \\
\\
Baselines & Refers to reference points or initial measurements that serve as a starting point for comparison or evaluation. Baselines provide a foundation for assessing the performance, effectiveness, or efficiency of systems, algorithms, models, or solutions. \\
\\
Assumptions & Refers to the statements or conditions that are considered to be true or valid for the purpose of designing, developing, or analysing systems, algorithms, models, or solutions. Assumptions simplify problem-solving processes by providing a set of predefined conditions or constraints under which a particular approach or solution is expected to work correctly. \\
\\
Examples & Refers to specific instances or data points that are used to illustrate or demonstrate a concept, principle, or the behavior of an algorithm or model. These examples serve to showcase the application of a technique or highlight particular characteristics, allowing for a clearer understanding and communication of the ideas involved. \\
\\
Empirical Findings & Refers to observations, data, or evidence obtained through direct observations, experiments, or measurements in the real world. They are based on empirical evidence rather than theoretical or speculative reasoning. \\
\\
Justifications & Refers to present reasons, evidence, or logical justifications in support of a particular claim, position, or viewpoint. It involves making a persuasive case or engaging in a reasoned debate. \\
\\
Challenges & Refers to difficulties, obstacles, or problems that need to be addressed or overcome in a particular context or task. They may arise due to technical, theoretical, practical, or ethical factors. \\
\\
Reproducibility & Refers to the ability to reliably recreate the same results or outputs from a given model or experiment, given the same input data and configuration settings, by providing the complete source code and using openly available tools and datasets. \\
\\
Motivations & Refers to the reasons, goals, or driving factors behind a particular study, project, or research endeavor. A gap refers to a missing or unaddressed aspect or area within existing knowledge or literature, which motivates further investigation or research. \\
\bottomrule
\end{tabular}
\caption{GPT-4 generated definitions for rigour criteria.}\label{tab:rigour_definitions}
\end{table*}

\subsection{Sentence labels}\label{tab:breakdown_labels}
\begin{table}[H]
\centering
\begin{tabular}{lc}
\toprule
\textbf{Criterion} & \textbf{Frequency} \\
\midrule
Settings & 251,755 \\
Benchmarks & 40,402 \\
Baselines & 30,100 \\
Generalisation & 20,527 \\
Reproducibility & 13,185 \\
Motivations & 11,820 \\
Biases & 10,549 \\
Assumptions & 5,726 \\
Robustness & 5,490 \\
Examples & 3,513 \\
Empirical Findings & 2,994 \\
Limitations & 2,048 \\
Justification & 1,053 \\
Constraints & 990 \\
\midrule
Total & 400,152 \\
\bottomrule
\end{tabular}
\caption{Breakdown of sentence labels for REF, ACL and ICLR papers.}
\end{table}

\subsection{4* vs non-4* rigour examples}\label{rigour_examples}
Examples comparing 4* and non-4* sentences, shown in Table \ref{compare_rigour}.

% \noindent\textbf{Reproducibility:}
% \begin{itemize}
%     \item 4*: We validate our findings through numerical experiments where our theory accurately predicts empirical findings and remains consistent with observations in deep neural networks.
%     \item non-4*: Our framework provides a reproducible and easy-to-use entry point for the development and evaluation of future bias mitigation algorithms in deep learning.
% \end{itemize}

% \noindent\textbf{Benchmark:}

% \begin{itemize}
%     \item 4*: We design experiments to thoroughly test and objectively score metrics on their ability to measure the diversity and fidelity of generated graphs, as well as their sample and computational efficiency.
%     \item non-4*: Ten years later, the ImageNet dataset is still one of the main benchmarks for state-of-the-art computer vision models (Krizhevsky et al., 2012; Simonyan & Zisserman, 2015; He et al., 2016; Liu et al., 2018; Howard et al., 2019; Touvron et al., 2021; Radford et al., 2021).
% \end{itemize}

% \noindent\textbf{Assumption:}

% \begin{itemize}
%     \item 4*: A prerequisite to comparing a machine's performance to human intelligence is, hence, the verification that machines can exhibit a sensitivity to context that would allow them to perform as well on cases that require reasoning about exceptions as on cases that require recalling generic associations.
%     \item non-4*: Such a preference-based decision procedure would then allow stronger valued evidence to override weaker one.
% \end{itemize}

\begin{table*}[h]
\centering
\resizebox{\textwidth}{!}{
\begin{tabular}{lp{0.4\textwidth}p{0.4\textwidth}}
\toprule
\textbf{Criterion} & \textbf{4*} & \textbf{non-4*} \\ 
\midrule
Reproducibility & 
We validate our findings through numerical experiments where our theory accurately predicts empirical findings and remains consistent with observations in deep neural networks. & 
Our framework provides a reproducible and easy-to-use entry point for the development and evaluation of future bias mitigation algorithms in deep learning. \\ 
\addlinespace
\addlinespace

Benchmark & 
We design experiments to thoroughly test and objectively score metrics on their ability to measure the diversity and fidelity of generated graphs, as well as their sample and computational efficiency. & 
Ten years later, the ImageNet dataset is still one of the main benchmarks for state-of-the-art computer vision models (Krizhevsky et al., 2012; Simonyan \& Zisserman, 2015; He et al., 2016; Liu et al., 2018; Howard et al., 2019; Touvron et al., 2021; Radford et al., 2021). \\
\addlinespace
\addlinespace

Assumption & 
A prerequisite to comparing a machine's performance to human intelligence is, hence, the verification that machines can exhibit a sensitivity to context that would allow them to perform as well on cases that require reasoning about exceptions as on cases that require recalling generic associations. &  
Such a preference-based decision procedure would then allow stronger valued evidence to override weaker one. \\ 
\bottomrule
\end{tabular}}
\caption{Comparison of 4* and non-4* sentences from different rigour criteria.}\label{compare_rigour}
\end{table*}

\subsection{Appending vs individual}\label{fig:query}
\begin{figure*}[!p]
\centering
\includegraphics[width=1\textwidth]{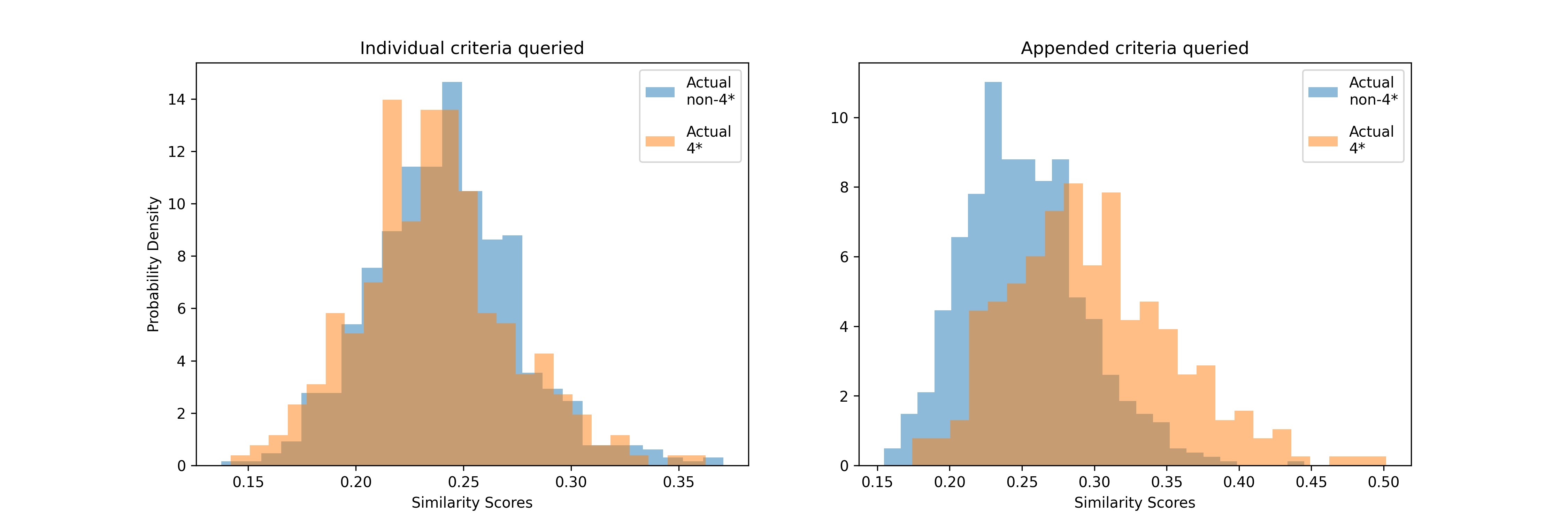}
\caption{Comparing the difference when the criteria are appended together (Best combination on the right) vs individual criterion queries and then adding the similarity score (Best criteria added together on the left)}\label{appended_vs_individual}
\end{figure*}

Appending criteria together and querying documents resulted in better separation, as shown in Table \ref{appended_vs_individual}.

\subsection{Certainty-aspect breakdown}\label{app:aspect_breakdown}
Full breakdown of certainity-aspect for each rigour criterion. As shown in Table \ref{tab:certainty_table} and Table \ref{tab:uncertainty_table}.

\begin{table*}[!hp]
\centering
\resizebox{\textwidth}{!}{
\begin{tabular}{lcccccccccc}
\toprule
\textbf{Criterion} & \textbf{Framing} & \textbf{Suggestion} & \textbf{Extent} & \textbf{Condition} & \textbf{Probability} & \textbf{Number} \\
\midrule
Settings & 1.14 & -0.37 & 0.07 & 0.47 & -1.83 & -0.55\\
Benchmarks & 2.09 & -3.71 & 0.87 & -0.44 & 1.14 & 0.53\\
Biases & 3.60 & -1.11 & -0.61 & 0.05 & 0.76 & -0.21\\
Robustness & 4.82 & -3.09 & -0.17 & -0.21 & 0.93 & -0.86\\
Reproducibility & 1.50 & -1.08 & 0.32 & 0.03 & -0.61 & -0.49\\
Constraints & 2.12 & -2.56 & 0.72 & -2.61 & 5.38 & 0.70\\
Baselines & 0.93 & -0.70 & 0.08 & 0.02 & 1.69 & -0.45\\
Limitations & 2.75 & 0.11 & -0.46 & -0.37 & -2.80 & 0.05\\
Generalisation & 1.39 & -0.36 & -0.12 & 0.17 & 0.13 & -0.09\\
Motivations & -1.84 & -2.27 & -0.10 & 0.62 & -0.38 & -0.32\\
Empirical Findings & -0.93 & -0.08 & -0.03 & 0.02 & 1.74 & 0.02\\
Assumptions & 1.06 & -1.65 & 0.01 & 1.04 & -0.24 & -0.34\\
Examples & 0.74 & 0.13 & 0.15 & -1.44 & 2.68 & 2.99\\
Justification & 0.53 & -1.42 & -0.28 & -0.13 & -1.56 & 2.53\\
\bottomrule
\end{tabular}}
\caption{Certainty Breakdown}\label{tab:certainty_table}
\end{table*}

\begin{table*}[!hp]
\centering
\resizebox{\textwidth}{!}{
\begin{tabular}{lcccccccc}
\toprule
\textbf{Criterion} & \textbf{Framing} & \textbf{Suggestion} & \textbf{Extent} & \textbf{Condition} & \textbf{Probability} & \textbf{Number} \\
\midrule
Settings & 0.24 & 0.37 & 0.31 & 0.03 & 0.31 & -0.20\\
Benchmarks & 0.04 & -0.69 & 1.48 & -0.06 & -1.21 & -0.03\\
Biases & -0.26 & -0.21 & -0.23 & 0.03 & -1.72 & -0.09\\
Robustness & 0.05 & -0.09 & 0.25 & -0.03 & -1.15 & -0.44\\
Reproducibility & 0.16 & -0.27 & 0.38 & 0.02 & 0.03 & 0.02\\
Constraints & -0.99 & -0.84 & 1.50 & 0.02 & -3.26 & -0.18\\
Baselines & -0.17 & -0.40 & 0.35 & 0.03 & -1.21 & -0.15\\
Limitations & 1.59 & 0.17 & -1.43 & 0.00 & 0.65 & -0.27\\
Generalisation & 0.65 & 0.08 & -0.55 & 0.00 & -1.22 & -0.07\\
Motivations & 1.78 & 0.05 & 0.23 & 0.07 & 1.82 & 0.33\\
Empirical Findings & 0.26 & -0.02 & -1.45 & 0.00 & 0.38 & 0.09\\
Assumptions & 0.17 & -0.10 & 0.47 & -0.06 & -0.24 & -0.12\\
Examples & 0.36 & -0.03 & -2.81 & 0.00 & -2.38 & -0.37\\
Justification & 0.51 & 0.65 & -2.72 & 0.00 & 2.19 & -0.29\\
\bottomrule
\end{tabular}}
\caption{Uncertainty Breakdown}\label{tab:uncertainty_table}
\end{table*}

\end{document}